\documentclass{article}
\usepackage[utf8]{inputenc}

\usepackage[margin=3.5cm]{geometry}
\usepackage{amsmath}
\usepackage{amsfonts}
\usepackage{amssymb}
\usepackage{array}
\usepackage{graphicx}
\usepackage{wrapfig}
\usepackage{hyperref}
\usepackage{authblk}
\usepackage{amsthm}
\usepackage{tikz-cd}
\usepackage{natbib}
\usepackage{gensymb}

\usepackage[utf8]{inputenc}
\usepackage[english]{babel}

\usepackage[framemethod=tikz]{mdframed}
\usetikzlibrary{shadows}

\everymath=\expandafter{\the\everymath\let\par\relax}
\everydisplay=\expandafter{\the\everydisplay\let\par\relax}

\newmdenv[shadow=true,shadowcolor=black,font=\sffamily,rightmargin=8pt]{shadedbox}

\newtheorem{theorem}{Theorem}[section]

\newtheorem{lemma}[theorem]{Lemma}

\newcommand{\id}{\mathrm{id}}

\newcommand{\R}{\mathbb{R}}

\newcommand{\SE}[1]{\ensuremath{\operatorname{SE}(#1)}}
\newcommand{\SO}[1]{\ensuremath{\operatorname{SO}(#1)}}

\newcommand{\ind}{\ensuremath{\operatorname{Ind}}}
\newcommand{\res}{\ensuremath{\operatorname{Res}}}
\DeclareMathOperator{\Hom}{Hom}

\newcommand{\h}[1]{\mathrm{h}_{#1}}

\title{Intertwiners between Induced Representations \\
\large with Applications to the Theory of Equivariant Neural Networks \\
(\emph{Preliminary Report})}
\date{February 2018}

\author[1]{Taco S. Cohen}
\author[2]{Mario Geiger}
\author[3]{Maurice Weiler}
\affil[1]{\small Qualcomm Research, Qualcomm Technologies Netherlands B.V.}
\affil[2]{\small EPFL}
\affil[3]{\small QUVA Lab, University of Amsterdam}

\usepackage{natbib}
\usepackage{graphicx}

\begin{document}

\maketitle

\begin{abstract}
    Group equivariant and steerable convolutional neural networks (regular and steerable G-CNNs)
    have recently emerged as a very effective model class for learning from signal data such as 2D and 3D images, video, and other data where symmetries are present.
    In geometrical terms, regular G-CNNs represent data in terms of \emph{scalar fields} (``feature channels''), whereas the steerable G-CNN can also use \emph{vector or tensor fields} (``capsules'') to represent data.
    In algebraic terms, the feature spaces in regular G-CNNs transform according to a \emph{regular representation} of the group $G$, whereas the feature spaces in Steerable G-CNNs transform according to the more general \emph{induced representations} of $G$.
    In order to make the network equivariant, each layer in a G-CNN is required to intertwine between the induced representations associated with its input and output space.

    In this paper we present a general mathematical framework for G-CNNs on homogeneous spaces like Euclidean space or the sphere.
    We show, using elementary methods, that the layers of an equivariant network are convolutional \emph{if and only if} the input and output feature spaces transform according to an induced representation.
    This result, which follows from G.W. Mackey's abstract theory on induced representations, establishes G-CNNs as a universal class of equivariant network architectures, and generalizes the important recent work of Kondor \& Trivedi on the intertwiners between regular representations.

    In order for a convolution layer to be equivariant, the filter kernel needs to satisfy certain linear equivariance constraints.
    The space of equivariant kernels has a rich and interesting structure, which we expose using direct calculations.

    Additionally, we show how this general understanding can be used to compute a basis for the space of equivariant filter kernels, thereby providing a straightforward path to the implementation of G-CNNs for a wide range of groups and manifolds.

\end{abstract}

\newpage
\section{Introduction}

In recent years, the Convolutional Neural Network (CNN) has emerged as the primary model class for learning from signals such as audio, images, and video.
Through the use of convolution layers, the CNN is able to exploit the spatial \emph{locality} of the input space, and the translational \emph{symmetry} (invariance) that is inherent in many learning problems.
Because convolutions are \emph{translation equivariant} (a shift of the input leads to a shift of the output), convolution layers preserve the translation symmetry.
This is important, because it means that further layers of the network can also exploit the symmetry.

Motivated by the success of CNNs, many researchers have worked on generalizations, 
leading to a growing body of work on \emph{group equivariant networks} \citep{Cohen2016-gm, Cohen2017-wl, Worrall2017-gl, Weiler2017-wc, Thomas2018-pd, Kondor2018-fm}.
Generalization has happened along two mostly orthogonal directions.
Firstly, the symmetry groups that can be exploited was expanded beyond pure translations, to other transformations such as rotations and reflections, by replacing convolutions with group convolutions \citep{Cohen2016-gm}.
The feature maps in these networks transform as scalar fields on the group $G$ or a homogeneous space $G/H$ \citep{Kondor_Trivedi_2018}.
We will refer to such networks as \emph{regular} G-CNNs, because the transformation law for scalar fields is known as the regular representation of $G$.

Initially, regular G-CNNs were implemented for planar images, acted on by discrete translations, rotations, and reflections.
Such discrete G-CNNs have the advantage that they are easy to implement, easy to use, fast, and result in improved results in a wide range of practical problems, making them a natural starting point for the generalization of CNNs.
However, the concept is much more general: because G-CNNs were formulated in abstract group theoretic language, they are easily generalized to any group or homogeneous space that we can sum or integrate over \citep{Kondor_Trivedi_2018}.
For instance, Spherical CNNs \citep{Cohen2018-sv} are G-CNNs for the 3D rotation group $G=\SO3$ acting on the sphere $S^2 = \SO3/\SO2$ (a homogeneous space for $\SO3$).

\begin{figure}[b]
    \centering
    \includegraphics[scale=0.8]{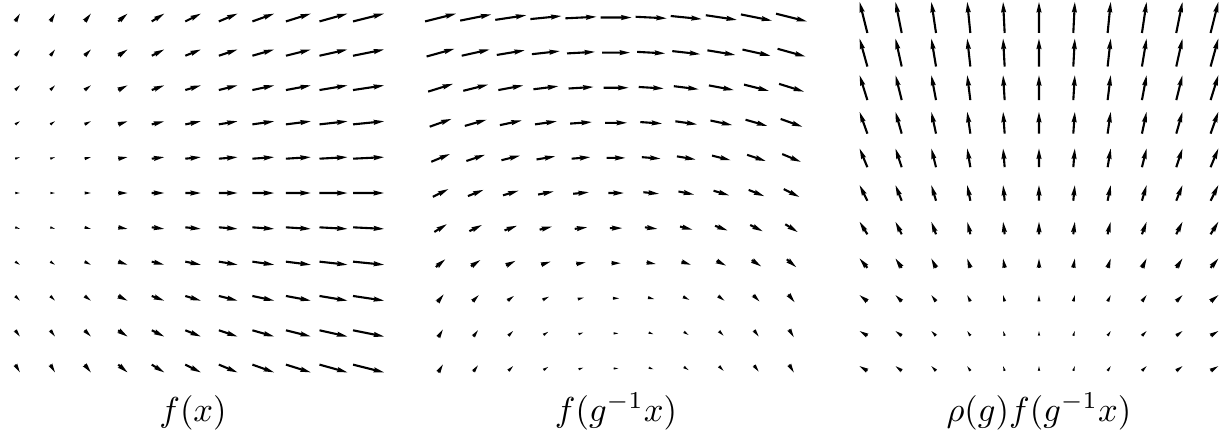}
    \caption{\small To transform a planar vector field by a $90\degree$ rotation $g$, first move each arrow to its new position, keeping its orientation the same, then rotate the vector itself. This is described by the induced representation $\pi = \ind_{\SO2}^{\SE2} \rho$, where $\rho(g)$ is a $2 \times 2$ rotation matrix that mixes the two coordinate channels.} \label{fig:rotate_field}
\end{figure}

The second direction of generalization corresponds to a move away from scalar fields.
Using connections to the theory of steerable filters \citep{Freeman1991-me} and induced representations \citep{ind_mackey2009, ind_fieldth, Gurarie1992-oj}, 
the feature space was generalized to vector- and tensor fields, and even more general spaces (sections of homogeneous vector bundles) \citep{Cohen2017-wl, Weiler2017-wc, Worrall2017-gl, Thomas2018-pd, Kondor2018-fm, Kondor2018-pp}.
We will refer to these networks as \emph{steerable} or \emph{induced} G-CNNs, because the filters in these networks are steerable, and the associated transformation law is called the induced representation (see Fig. \ref{fig:rotate_field}).

Thus, the general picture that has emerged is one of networks that use convolutions to map between spaces of sections of homogeneous vector bundles in a group equivariant manner.
The classical CNN, mapping scalar fields (a.k.a. feature channels) on the plane to scalar fields on the plane in a translation equivariant manner, is but one special case.
In this paper we study the general class of induced G-CNNs, and in particular the space of equivariant linear maps (intertwiners) between two induced representations associated with the input and output feature space of a network layer.
We show that any equivariant map between induced representations can be written as a (twisted) convolution / cross-correlation, thus generalizing the results of \cite{Kondor_Trivedi_2018}, who showed this for regular representations\footnote{Since the regular representation of $G$ on $G/H$ is the representation of $G$ induced from the trivial representation of $H$, the results of Kondor \& Trivedi can be obtained from ours by filling in the trivial representation $\rho(h) = 1$ whenever one encounters $\rho$ in this paper.}.

The induced representation has been studied extensively by physicists and mathematicians.
The word ``induced'' comes from the fact that the transformation law of e.g. a vector field can be inferred from the transformation law of an individual vector under the action of a certain isotropy (or ``stabilizer'') subgroup of the symmetry group.
For instance, when applying a 3D rotation $R \in \SO3$ to a vector field $f$ on the sphere, each vector $f(x)$ is moved to a new position $R^{-1} x \in S^2$ by the 3D rotation, and the vector itself is rotated in its tangent plane by a 2D rotation $r(R) \in \SO2$ (This is illustrated in Fig. \ref{fig:rotate_field} for a planar vector field).
Thus, we say that this vector field transforms according to the representation of $\SO3$ induced by the canonical representation of $\SO2$.
As another example, a higher order tensor transforms according to a different representation $\rho$ of $\SO2$, so a tensor \emph{field} on the sphere transforms according to a different induced representation of $\SO3$.

Induced representations are important in physics because they are the primary tool to construct irreducible representations, which enumerate the types of elementary particles of a physical (field) theory.
In representation learning, the idea of irreducible representations as elementary particles has been applied to formalize the idea of ``disentangling'' or ``capsules'' that represent distinct visual entities \citep{Cohen2014-hz, Cohen2015-pm}, each of which has a certain \emph{type} \citep{Cohen2017-wl}.
Indeed, we think of induced G-CNNs as the mathematically grounded version of Hinton's idea of capsules (sans dynamic routing, for now) \citep{Hinton2011-dw, Sabour2017-eh, Hinton2018-ck}.

The general formalism of fiber bundles has also been proposed as a geometrical tool for modelling early visual processing in the mammalian brain \citep{Petitot2003-tn}.
Although it is far too early to conclude anything, this convergence of physics, neuroscience, and machine learning suggests that field theories are not just for physicists, but provide a generally useful model class for natural and man-made learning systems.

\subsection{Outline and Summary of Results}

In order to understand and properly define the induced representation, we need some notions from group- and representation theory, such as groups, cosets, double cosets, quotients, sections, and representations.
In section \ref{sec:math_background} we will define these concepts and illustrate them with two examples: the rotation group $\SO3$ and Euclidean motion group $\SE3$.
Although necessary for a detailed understanding of the rest of the paper, this section is rather dry and may be skimmed on a first reading.

Induced representations are defined in section \ref{sec:induced_reps}.
We present two of the many equivalent realizations of the induced representation.
The first realization describes the transformation law for the vector space $\mathcal{I}_C$ of sections of a vector bundle over $G/H$, such as vector fields over the sphere $S^2 = \SO3 / \SO2$.
This realization is geometrically natural, and such vector fields can be stored efficiently in computer memory, making them the preferred realization for implementations of induced G-CNNs.
The downside of this realization is that, due to the use of an arbitrary frame of reference (choice of section), the equations describing it get quite cumbersome.
For this reason, we also discuss the induced representation realized in the space $\mathcal{I}_G$ of vector-valued functions on $G$, having a certain kind of symmetry (the space of Mackey functions).
We define a ``lifting'' isomorphism from $\mathcal{I}_C$ to $\mathcal{I}_G$ to show that they are equivalent:
\begin{equation}
    \begin{tikzcd}
          \mathcal{I}_{C} \arrow[r, "\Lambda"] & \mathcal{I}_{G}
    \end{tikzcd}
\end{equation}

In section \ref{sec:intertwiners_elementary} we study the space of linear equivariant maps, or intertwiners, between two representations $\pi_1$ and $\pi_2$, induced from representations of subgroups $H_1 \leq G$ and $H_2 \leq G$.
Denoting this space by $\mathcal{H}_G$ or $\mathcal{H}_{C}$ (depending on the chosen realization of $\mathcal{I}_G$ or $\mathcal{I}_C$), we find that (of course) they are equivalent, and more importantly, that any equivariant map $f \in \mathcal{H}_*$ can be written as a special kind of convolution or correlation with an equivariant kernel $\kappa$ on $G$ or $G/H_1$, respectively.
Furthermore, these spaces of equivariant kernels, denoted $\mathcal{K}_G$ and $\mathcal{K}_C$, are shown to be equivalent to a space of kernels on the double coset space $H_2 \backslash G / H_1$, denoted $\mathcal{K}_D$.
This is summarized in the following diagram of isomorphisms:
\begin{equation}
    \begin{tikzcd}
        & \mathcal{H}_{C} \arrow[r, "\Lambda_\mathcal{H}"] & \mathcal{H}_G \\
        \mathcal{K}_D     \arrow[r, "\Omega_{\mathcal{K}}"] 
        & \mathcal{K}_{C} \arrow[u, "\Gamma_C"] \arrow[r, "\Lambda_{\mathcal{K}}"] 
        & \mathcal{K}_G \arrow[u, "\Gamma_G"]
    \end{tikzcd}
\end{equation}
The map $\Gamma_G$ takes a kernel $\kappa \in \mathcal{K}_G$ to the ``neural network layer'' $\kappa \, \star \in \mathcal{H}_G$,
\begin{equation}
    \kappa \star : \mathcal{I}_G \rightarrow \mathcal{I}_G,
\end{equation}
by using the kernel in a cross-correlation denoted $\star$. The map $\Gamma_C$ is defined similarly.
That $\Gamma_*$ is an isomorphism means that any equivariant map $\Phi \in \mathcal{H}_*$ can be written as a convolution with an appropriate kernel $\kappa \in \mathcal{K}_*$.

The kernels in $\mathcal{K}_C$ and $\mathcal{K}_G$ have to satisfy certain equivariance constraints.
These constraints can be largely resolved by moving to $\mathcal{K}_D$, where finding a solution is typically easier.
Using the results of this paper, finding a basis for the space of equivariant filters for a new group should be relatively straightforward.

Having seen the main results derived in a relatively concrete manner, we proceed in section \ref{sec:intertwiners_mackey} to show how these results relate to Mackey's theory of induced representations, which is usually presented in a more abstract language.
Then, in section \ref{sec:examples}, we show how to actually compute a basis for $\mathcal{K}_C$ for the case of $G = \SO3$ and $G = \SE3$.

\newpage
\section{Mathematical Background}
\subsection{General facts about Groups and Quotients}
\label{sec:math_background}

\begin{figure}
    \centering
    \includegraphics{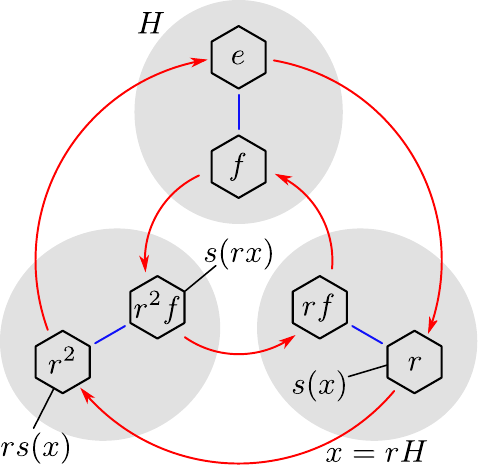}
    \caption{A Cayley diagram of the group $D3$ of symmetries of a triangle. The group is generated by rotations $r$ and flips $f$. The elements of the group are indicated by hexagons. The red arrows correspond to right multiplication by $r$, while the blue lines correspond to right multiplication by $f$. Cosets of the group of flips ($H = \{e, f\}$) are shaded in gray. As always, the cosets partition the group. As coset representatives, we choose $s(H) = e$, $s(rH) = r$, and $s(r^2 H) = r^2f$. The difference between $s(rx)$ and $r s(x)$ is indicated. For this choice of section, we must set $\h{}(x, r) = \h{}(rH, r) = f$, so that $s(r x) \h{}(x, r) = (r^2 f) (f) = r^2 = r s(x)$.}
    \label{fig:cosets_hfunc}
\end{figure}

Let $G$ be a group and $H$ a subgroup of $G$.
A left coset of $H$ in $G$ is a set $g H = \{ gh \; | \; h \in H \}$ for $g \in G$.
The cosets form a partition of $G$.
The set of all cosets is called the quotient space or coset space, and is denoted $G/H$.
There is a canonical projection $p : G \rightarrow G/H$ that assigns to each element $g$ the coset it is in.
This can be written as $p(g) = gH$.
Fig. \ref{fig:cosets_hfunc} provides an illustration for the group of symmetries of a triangle, and the subgroup $H$ of reflections.

The quotient space carries a left action of $G$, which we denote with $ux$ for $u \in G$ and $x \in G/H$.
This works fine because this action is associative with the group operation: 
\begin{equation}
    \label{eq:proj_assoc}
    u (gH) = (ug) H.
\end{equation}
for $u, g \in G$. One may verify that this action is well defined, i.e. does not depend on the particular coset representative $g$.
Furthermore, the action is transitive, meaning that we can reach any coset from any other coset by transforming it with an appropriate $u \in G$.
A space like $G/H$ on which $G$ acts transitively is called a homogeneous space for $G$.
Indeed, any homogeneous space is isomorphic to some quotient space $G/H$.

A section of $p$ is a map $s : G/H \rightarrow G$ such that $p \circ s = \id_{G/H}$.
We can think of $s$ as choosing a coset representative for each coset, i.e. $s(x) \in x$.
In general, although $p$ is unique, $s$ is not; there can be many ways to choose coset representatives.
However, the constructions we consider will always be independent of the particular choice of section.

Although it is not strictly necessary, we will assume that $s$ maps the coset $H=eH$ of the identity to the identity $e \in G$:
\begin{equation}
    s(H) = e
\end{equation}
We can always do this, for given a section $s'$ with $s'(H) = h \neq e$, we can define the section $s(x) = h^{-1} s'(h x)$ so that $s(H) = h^{-1} s'(hH) = h^{-1} s'(H) = h^{-1} h = e$.
This is indeed a section, for $p(s(x)) = p(h^{-1} s'(hx)) = h^{-1} p(s'(hx)) = h^{-1} h x = x$ (where we used Eq. \ref{eq:proj_assoc} which can be rewritten as $u p(g) = p(ug)$).

One useful rule of calculation is
\begin{equation}
    (g s(x)) H = g (s(x) H) = g x = s(g x) H,
\end{equation}
for $g \in G$ and $x \in G/H$.
The projection onto $H$ is necessary, for in general $gs(x) \neq s(gx)$.
These two terms are however related, through a function $\h{} : G/H \times G \rightarrow H$, defined as follows:
\begin{equation}
    g s(x) = s(g x) \h{}(x, g)
\end{equation}
That is,
\begin{equation}
    \label{eq:h_def}
    \boxed{
    \h{}(x, g) = s(g x)^{-1} g s(x)
    }.
\end{equation}
We can think of $\h{}(x, g)$ as the element of $H$ that we can apply to $s(g x)$ (on the right) to get $g s(x)$.
The $\h{}$ function will play an important role in the definition of the induced representation, and is illustrated in Fig. \ref{fig:cosets_hfunc}.

From the fiber bundle perspective, we can interpret Eq. \ref{eq:h_def} as follows.
The group $G$ can be viewed as a principal bundle with base space $G/H$ and fibers $gH$.
If we apply $g$ to the coset representative $s(x)$, we move to a different coset, namely the one represented by $s(gx)$ (representing a different point in the base space).
Additionally, the fiber is twisted by the right action of $\h{}(x, g)$.
That is, $\h{}(x, g)$ moves $s(gx)$ to another element in its coset, namely to $g s(x)$.

The following composition rule for $\h{}$ is very useful in derivations:
\begin{equation}
    \label{eq:h_g1g2}
    \begin{aligned}
        \h{}(x, g_1 g_2)
        &= s(g_1 g_2 x)^{-1} g_1 g_2 s(x) \\
        &= [s(g_1 g_2 x)^{-1} g_1 s(g_2 x)] [s(g_2 x)^{-1} g_2 s(x)] \\
        &= \h{}(g_2 x, g_1) \h{}(x, g_2)
    \end{aligned}
\end{equation}
For elements $h \in H$, we find:
\begin{equation}
    \h{}(H, h) = s(H)^{-1} h s(H) = h.
\end{equation}
Also, for any coset $x$,
\begin{equation}
    \label{eq:h_H_sx}
    \h{}(H, s(x)) = s(s(x) H)^{-1} s(x) s(H) =
    s(H) = e.
\end{equation}
Using Eq. \ref{eq:h_g1g2} and \ref{eq:h_H_sx}, this yields,
\begin{equation}
    \h{}(H, s(x) h) = \h{}(h H, s(x)) \h{}(H, h) = h,
\end{equation}
for any $h \in H$ and $x \in G/H$.

For $x = H$, Eq. \ref{eq:h_def} specializes to:
\begin{equation} \label{eq:h1_def}

    g = g s(H) = s(gH) \h{}(H, g) \equiv s(gH) \h{}(g),

\end{equation}
where we defined
\begin{equation}

    \boxed{
    \h{}(g) = \h{}(H, g) = s(gH)^{-1} g
    }
\end{equation}
This shows that we can always factorize $g$ \textit{uniquely} into a part $s(gH)$ that represents the coset of $g$, and a part $\h{}(g) \in H$ that tells us where $g$ is within the coset:
\begin{equation}\label{eq:unique_decomposition}
    g = s(gH) \h{}(g)
\end{equation}
A useful property of $\h{}(g)$ is that for any $h \in H$,
\begin{equation}
    \label{eq:h_gh}
    \h{}(g h) = s(g h H)^{-1} g h = s(gH)^{-1} g h = \h{}(g) h.
\end{equation}
It is also easy to see that
\begin{equation}
    \label{eq:hsx_is_e}
    \h{}(s(x)) = e.
\end{equation}

When dealing with different subgroups $H_1$ and $H_2$ of $G$ (associated with the input and output space of an intertwiner), we will write $h_i$ for an element of $H_i$, $s_i : G/H_i \rightarrow G$, for the corresponding section, and $\h{i} : G/H_i \times G \rightarrow H_i$ for the $\h{}$-function (for $i=1,2$).

\subsection{Double cosets}
\label{sec:double_cosets}

A $(H_2,H_1)$-double coset is a set of the form $H_2gH_1$ for $H_2, H_1$ subgroups of $G$.
The space of $(H_2,H_1)$-double cosets is called $H_2\backslash G / H_1 \equiv \{H_2gH_1 \,|\, g\in G\}$.
As with left cosets, we assume a section $\gamma:H_2\backslash G / H_1\to G$ is given, satisfying $\gamma(H_2gH_1) \in H_2gH_1$.

The double coset space $H_2\backslash G/H_1$ can be understood as the space of $H_2$-orbits in $G/H_1$, that is, $H_2\backslash G/H_1 = \{ H_2 x | x \in G/H_1 \}.$
Note that although $G$ acts transitively on $G/H_1$ (meaning that there is only one $G$-orbit in $G/H_1$), the subgroup $H_2$ does not.
Hence, the space $G/H_1$ splits into a number of disjoint orbits $H_2x$ (for $x = gH_1 \in G/H_1$), and these are precisely the double cosets $H_2gH_1$.

Of course, $H_2$ \emph{does} act transitively within a single orbit $H_2x$, sending $x \mapsto h_2x$ (both of which are in $H_2x$, for $x \in G/H_1$).
In general this action is not necessarily fixed point free which means that there may exist some $h_2 \in H_2$ which map the left cosets to themselves.
These are exactly the elements in the stabilizer of $x=gH_1,$ given by
\begin{equation}
    \label{eq:coset_stabilizer}
    \begin{aligned}
        H_2^x &= \{h \in H_2 \, | \, h x = x \} \\
              &= \{h \in H_2 \, | \, h s_1(x) H_1 = s_1(x) H_1 \} \\
              &= \{h \in H_2 \, | \, h s_1(x) \in s_1(x)H_1 \} \\
              &= \{h \in H_2 \, | \, h \in s_1(x)H_1 s_1(x)^{-1} \} \\
              &= s_1(x) H_1 s_1(x)^{-1} \cap H_2.
    \end{aligned}
\end{equation}
Clearly, $H_2^x$ is a subgroup of $H_2$.
Furthermore, $H_2^x$ is conjugate to (and hence isomorphic to) the subgroup $s_1(x)^{-1} H_2^x s_1(x) = H_1 \cap s_1(x)^{-1} H_2 s_1(x)$, which is a subgroup of $H_1$.

For double cosets $x \in H_2\backslash G /H_1$, we will overload the notation to $H^x_2 \equiv H^{\gamma(x) H_1}_2$.
Like the coset stabilizer, this double coset stabilizer can be expressed as
\begin{equation}
    \label{eq:double_coset_stabilizer}
    H_2^x = \gamma(x) H_1 \gamma(x)^{-1} \cap H_2
\end{equation}

\subsection{Semidirect products}

For a semidirect product group $G$, such as $\SE2 = \R^2 \rtimes \SO2$, some things simplify. Let $G=N \rtimes H$ where $H \leq G$ is a subgroup, $N \leq G$ is a normal subgroup and $N \cap H = \{e\}$.
For every $g \in G$ there is a unique way of decomposing it into $nh$ where $n \in N$ and $h \in H$.
Thus, the left $H$ coset of $g \in G$ depends only on the $N$ part of $g$:
\begin{equation}
    gH = nhH = nH
\end{equation}
It follows that for a semidirect product group, we can define the section so that it always outputs an element of $N \subseteq G$, instead of a general element of $G$.
Specifically, we can set $s(gH) = s(nhH) = s(nH) = n$.
It follows that $s(n x) = n s(x) \;\; \forall n \in N, x \in G/H$.
This allow us to simplify expressions involving $\h{}$:
\begin{equation}
    \label{eq:semidirect_h}
    \begin{aligned}
    \h{}(x, g) 
    &= s(gx)^{-1} g s(x) \\
    &= s(g s(x) H)^{-1} g s(x) \\
    &= s(\underbrace{g s(x) g^{-1}}_{\in N} g H)^{-1} g s(x) \\
    &= \left( g s(x) g^{-1} \; s(gH) \right)^{-1} g s(x) \\
    &= s(gH)^{-1} g \\
    &= \h{}(g) 
    \end{aligned}
\end{equation}

\subsection{Haar measure}

When we integrate over a group $G$, we will use the Haar measure, which is the essentially unique measure $dg$ that is invariant in the following sense:
\begin{equation}
    \int_G f(g) dg = \int_G f(ug) dg \quad \forall u \in G.
\end{equation}
Such measures always exist for locally compact groups, thus covering most cases of interest \citep{Folland1995-dg}.
For discrete groups, the Haar measure is the counting measure, and integration can be understood as a discrete sum.

We can integrate over $G/H$ by using an integral over $G$,
\begin{equation}
    \int_{G/H} f(x) dx = \int_G f(gH) dg.
\end{equation}

\newpage
\section{Induced Representations}
\label{sec:induced_reps}

The induced representation can be realized in various equivalent ways.
We first discuss the realization through Mackey functions, which is easier to deal with mathematically.
We then discuss the realization on functions\footnote{Technically, we should work with sections of a homogeneous vector bundle $E \rightarrow G/H$, instead of functions, but to keep things simple we will not. This is not a problem as long as one can find a continuous section $G/H \rightarrow G$ of the bundle $G \rightarrow G/H$ that is defined almost everywhere.} on $G/H$, which gives more complicated equations, but which can be implemented more efficiently in software, and more clearly conveys the geometrical meaning.

\subsection{Realization through Mackey Functions}

The space of Mackey functions for a representation $(\rho, V)$ of $H \leq G$ is defined as:
\begin{equation}
    \label{eq:mackey_funcs}
    \mathcal{I}_G = 
    \{f : G \rightarrow V \; | \; f(gh) = \rho(h^{-1}) f(g), \; \forall g \in G, h \in H\}
\end{equation}
One may verify that this is a vector space.
The induced representation $\pi = \ind_H^G \rho$ acting on $\mathcal{I}_G = \ind_H^G V$ is defined as:
\begin{equation}
    \label{eq:def_ind_G}
    [\pi(u) f](g) = f(u^{-1} g),
\end{equation}
for $f \in \mathcal{I}_G$ and $u, g \in G$.
It is clear that $\pi(u) f$ is in $\mathcal{I}_G$ (i.e. satisfies the equivariance condition in Eq. \ref{eq:mackey_funcs}), because the left and right action commute.

\subsection{Realization through Functions on $G/H$}

Another way to realize the induced representation is on the space of vector valued functions on the quotient $G/H$,
\begin{equation}
    \label{eq:def_I_C}
    \mathcal{I}_{C} = \{ f: G/H \rightarrow V \}.
\end{equation}
Using a section $s : G/H \rightarrow G$ of the canonical projection $p : G \rightarrow G/H$, and the function $\h{} : G/H \times G \rightarrow H$ (Eq. \ref{eq:h_def}), we can define $\pi = \ind_H^G \rho$ as:
\begin{equation}
    \label{eq:def_ind_C}
    \begin{aligned}
        \lbrack \pi(u) f](x) &= \rho(\h{}(x, u^{-1})^{-1}) f(u^{-1} x) \\
                      &= \rho(\h{}(u^{-1} x, u)) f(u^{-1} x). 
    \end{aligned}
\end{equation}
The meaning of this equation is that to transform the function $f : G/H \rightarrow V$, we have to do two things: first, we take each vector $f(u^{-1} x) \in V$ and attach it at position $x$ of the transformed function $\pi(u) f$, without changing it.
Secondly, we need to transform the vector itself by the representation $\rho$ of $H$.
This is demonstrated in Fig. \ref{fig:rotate_field}.

To see that Eq. \ref{eq:def_ind_C} does indeed define a representation, we expand the definition of $\pi$ twice and use the composition rule for $\h{}$ (Eq. \ref{eq:h_g1g2}):
\begin{equation}
    \begin{aligned}
        \lbrack \pi(u) [\pi(v) f]](x)
        &=
        \rho(\h{}(x, u^{-1})^{-1}) [\pi(v) f](u^{-1} x) \\
        &= 
        \rho(\h{}(x, u^{-1})^{-1}) \rho(\h{}(u^{-1} x, v^{-1})^{-1}) f(v^{-1} u^{-1} x) \\
        &=
        \rho((\h{}(u^{-1} x, v^{-1}) \h{}(x, u^{-1}))^{-1}) f((uv)^{-1} x) \\
        &=
        \rho((\h{}(x, v^{-1} u^{-1}) f((uv)^{-1} x) \\
        &=
        \rho((\h{}(x, (uv)^{-1}) f((uv)^{-1} x) \\
        &=
        [\pi(uv) f](x)
    \end{aligned}
\end{equation}

\subsection{Equivalence of the Realizations}

To show that the constructions are equivalent, we will define a lifting map $\Lambda$ of functions $f : G/H \rightarrow V$ (i.e. $f \in \mathcal{I}_{C}$) to Mackey functions $f : G \rightarrow V$ (i.e. $f \in \mathcal{I}_G$), and show that it is a bijection that commutes with the two definitions of $\ind$.

The lift $\Lambda : \mathcal{I}_{C} \rightarrow \mathcal{I}_G$ and its inverse are defined as:
\begin{align}
    [\Lambda f](g)      &= \rho(\h{}(g)^{-1}) f(gH),\\
    [\Lambda^{-1} f'](x) &= f'(s(x)).
\end{align}
for $f \in \mathcal{I}_{C}$ and $f' \in \mathcal{I}_G$.
The idea behind this definition is that a Mackey function $f' \in \mathcal{I}_G$ is determined by its value on coset representatives $s(x)$, because by Eq. \ref{eq:unique_decomposition} and \ref{eq:mackey_funcs} it satisfies $f'(g) = f'(s(x) h) = \rho(h) f'(s(x))$.
Hence, setting $f(x) = f'(s(x))$ does not lose information.
Specifically, we can reconstruct $f'$ by setting $f'(g) = \rho(\h{}(g)^{-1}) f(gH)$.

It is easy to show, using Eq. \ref{eq:h_gh}, that $\Lambda f$ satisfies the equivariance condition (Eq. \ref{eq:mackey_funcs}):
\begin{equation}
    [\Lambda f](gh) = \rho(\h{}(gh)^{-1}) \; f(ghH) = \rho(h^{-1} \h{}(g)^{-1}) \; f(gH) = \rho(\h{}^{-1}) [\Lambda f](g)
\end{equation}
So indeed $\Lambda f \in \mathcal{I}_G$ for $f \in \mathcal{I}_{C}$.
To verify that $\Lambda$ is inverse to $\Lambda^{-1}$, 
use Eq. \ref{eq:h1_def}:
\begin{equation}

        \lbrack\Lambda[\Lambda^{-1} f]](g) 

        = \rho(\h{}(g)^{-1}) \, f(s(gH)) \\
        = f(s(gH) \h{}(g)) \\
        = f(g).

\end{equation}
For the opposite direction, using $\h{}(s(x)) = e$,
\begin{equation}
    [\Lambda^{-1}[\Lambda f]](x) = [\Lambda f](s(x)) = \rho(\h{}(s(x))^{-1}) \; f(s(x) H) = f(x).
\end{equation}

Finally, we show that $\Lambda$ commutes with the two definitions of the induced representation (Eq. \ref{eq:def_ind_G} and \ref{eq:def_ind_C}).
Let $\pi$ be the induced representation on $\mathcal{I}_G$ and $\pi'$ the induced rep on $\mathcal{I}_{C}$.
For $f \in \mathcal{I}_{C}$,
\begin{equation}
    \begin{aligned}
        \lbrack\Lambda[\pi'(u) f]](g)
        &= \rho(\h{}(g)^{-1})             \quad [\pi'(u) f](gH) \\
        &= \rho(\h{}(g)^{-1}) \; \rho(\h{}(gH, u^{-1})^{-1})  \quad f(u^{-1} gH) \\
        &= \rho((\h{}(gH, u^{-1}) \h{}(g))^{-1})              \quad f(u^{-1} gH) \\
        &= \rho(\h{}(H, u^{-1} g)^{-1})                    \quad f(u^{-1} gH) \\
        &= \rho(\h{}(u^{-1} g)^{-1}) \;                    \quad f(u^{-1} gH) \\
        &= [\Lambda f](u^{-1} g) \\
        &= [\pi(u) [\Lambda f]](g) \\
    \end{aligned}
\end{equation}
It follows that $(\pi, \, \mathcal{I}_G)$ and $(\pi', \, \mathcal{I}_C)$ are isomorphic representations of $G$.

\subsection{Some basic properties of induction}

We state some basic facts about induced representations.
Proofs can be found in \cite{ind_mackey2009}.

\begin{theorem}[Induction in stages]
    Let $G$ be a group and $K \leq H \leq G$ subgroups of $G$, and let $(\rho, V)$ be a representation of $K$, then:
    \begin{equation}
        \ind_H^G \ind_K^H \rho \simeq \ind_K^G \rho.
    \end{equation}
\end{theorem}

\begin{theorem}
    The induced representation of a direct sum of representations is the direct sum of the induced representations:
    \begin{equation}
        \ind_H^G \bigoplus_i \rho_i \simeq \bigoplus_i \ind_H^G \rho_i. 
    \end{equation}
\end{theorem}

\newpage
\section{Intertwiners: Elementary Approach}
\label{sec:intertwiners_elementary}

We would like to understand the structure of the space of intertwiners between two induced representations $(\pi_1, \, \mathcal{I}_G^1)$ and $(\pi_2, \, \mathcal{I}_G^2)$:
\begin{equation}
    \mathcal{H}_G = \Hom_G(\mathcal{I}_G^1, \, \mathcal{I}_G^2) = \{ \Phi : \mathcal{I}^1_G \rightarrow \mathcal{I}_G^2 \, | \, \Phi \pi_1(g) = \pi_2(g) \Phi, \;\;\; \forall g \in G \},
\end{equation}
and similarly for $\mathcal{H}_C = \Hom_G(\mathcal{I}_C^1, \, \mathcal{I}_C^2)$.

Using direct calculation, we will show that every map $\Phi$ in $\mathcal{H}_G$ or $\mathcal{H}_C$ can be written as a convolution or cross-correlation with an equivariant kernel.
We will start with the Mackey function approach.

\subsection{Intertwiners for $\mathcal{I}_G$}

Let $(\rho_1, V_1)$ be a representation of $H_1 \leq G$, and let $\pi_1 = \ind_{H_1}^G \rho_1$ be the induced representation acting on functions in $\mathcal{I}_G^1$.
Likewise, let $(\rho_2, V_2)$ be a representation of $H_2 \leq G$, and let $\pi_2 = \ind_{H_2}^G \rho_2$ be the induced representation acting on functions in $\mathcal{I}_G^2.$

A general linear map between vector spaces $\mathcal{I}_G^1$ and $\mathcal{I}_G^2$ can always be written as
\begin{equation}\label{eq:general_linear_mackey}
    [\kappa \cdot f](g) = \int_G \kappa(g, \, g') f(g') dg',
\end{equation}
using a two-argument operator-valued kernel $\kappa : G \times G \rightarrow \Hom(V_1, V_2)$.

In order for Eq. \ref{eq:general_linear_mackey} to define an equivariant map between $\mathcal{I}^1_G$ and $\mathcal{I}^2_G$, the kernel $\kappa$ must satisfy several constraints.
By (partially) resolving these constraints, we will show that Eq. \ref{eq:general_linear_mackey} can always be written as a cross-correlation, and that the space of admissible kernels is in one-to-one correspondence with the space of bi-invariant one-argument kernels $\mathcal{K}_C$, to be defined below.

\subsubsection{Equivariance $\Leftrightarrow$ Convolution}
\label{sec:equivar_biimpl_conv}
Since we are only interested in equivariant maps, we get a constraint on $\kappa$:
\begin{equation}
    \label{eq:two_arg_constraint_G}
    \begin{aligned}
        &&\lbrack \kappa \cdot [\pi_1(u) f]](g) 
        &\ =\ 
        \pi_2(u) [\kappa \cdot f](g) \\
        \Leftrightarrow\quad
        &&\int_G \kappa(g , g') f(u^{-1} g') dg'
        &\ =\ 
        \int_G \kappa(u^{-1} g, g') f(g') dg' \\
        \Leftrightarrow\quad
        &&\int_G \kappa(g , u g') f(g') dg'
        &\ =\ 
        \int_G \kappa(u^{-1} g, g') f(g') dg' \\
        \Leftrightarrow\quad
        &&\kappa(g, ug')
        &\ =\ 
        \kappa(u^{-1} g, g') \\
        \Leftrightarrow\quad
        &&\kappa(u g, u g')
        &\ =\ 
        \kappa(g, g')
    \end{aligned}
\end{equation}
Hence, without loss of generality, we can define the two-argument kernel $\kappa(\cdot, \cdot)$ in terms of a one-argument kernel:
\begin{equation}
    \kappa(g^{-1} g') \equiv \kappa(e, g^{-1} g') = \kappa(ge, gg^{-1}g') = \kappa(g, g').

\end{equation}
The application of $\kappa$ to $f$ reduces to a cross-correlation:
\begin{equation}
    [\kappa \star f](g) = \int_G \kappa(g^{-1} g') f(g') dg' = [\kappa \cdot f](g).
\end{equation}
It is also possible to define the one-argument kernel differently, so that we would get a convolution instead of a cross-correlation.

\subsubsection{Left equivariance of $\kappa$}
\label{sec:left_equivariance_M}

We want the result $\kappa \star f$ (or $\kappa \cdot f$) to live in $\mathcal{I}_G^2$, which means that this function has to satisfy the Mackey condition,
\begin{equation}
    \begin{aligned}
        &&\lbrack\kappa \star f](gh_2) &\ =\ \rho_2(h_2^{-1}) [\kappa \star f](g) \\
        \Leftrightarrow\quad
        &&\int_G \kappa((gh_2)^{-1} g') f(g') dg' &\ =\ \rho_2(h_2^{-1}) \int_G \kappa(g^{-1} g') f(g') dg' \\
        \Leftrightarrow\quad
        &&\kappa(h_2^{-1} g^{-1} g') &\ =\ \rho_2(h_2^{-1}) \kappa(g^{-1} g') \\
        \Leftrightarrow\quad
        &&\kappa(h_2 g) &\ =\ \rho_2(h_2) \kappa(g)
    \end{aligned}
\end{equation}
for all $h_2 \in H_2$ and $g \in G$.

\subsubsection{Right equivariance of $\kappa$}
\label{sec:right_equivariance_kappa}

The fact that $f \in \mathcal{I}^1_G$ satisfies the Mackey condition ($f(gh) = \rho_1(h) f(g)$ for $h \in H_1$) implies a symmetry in the correlation $\kappa \star f$.
That is, if we apply a right-$H_1$-shift to the kernel, i.e. $[R_{h} \kappa](g) = \kappa(gh)$, we find that
\begin{equation}
    \begin{aligned}
        \lbrack[R_{h} \kappa] \star f](g) &\ =\ \int_G \kappa(g^{-1} u h) f(u) du \\
                                      &\ =\ \int_G \kappa(g^{-1} u) f(u h^{-1}) du \\
                                      &\ =\ \int_G \kappa(g^{-1} u) \rho_1(h) f(u) du. \\

    \end{aligned}
\end{equation}
It follows that we can take (for $h \in H_1)$,
\begin{equation}
    \kappa(g h) = \kappa(g) \rho_1(h).
\end{equation}

\subsubsection{Resolving the right-equivariance constraint}

The above constraints show that the one-argument kernel $\kappa$ should live in the space of bi-equivariant kernels on $G$:
\begin{equation}
    \begin{aligned}
        \mathcal{K}_G = \{ \kappa : G \rightarrow \Hom(V_1, V_2) \, | \, 
        & \kappa(h_2 g h_1) = \rho_2(h_2) \kappa(g) \rho_1(h_1), \\
        & \forall g \in G, h_1 \in H_1, h_2 \in H_2 \}.
    \end{aligned}
\end{equation}
Here $\Hom(V_1, V_2)$ denotes the space of linear maps from $V_1$ to $V_2$.

We can resolve the right $H_1$-equivariance constraint $\kappa(gh_1) = \kappa(g) \rho_1(h_1)$ by defining $\kappa$ in terms of a kernel on the left coset space, i.e. $\overleftarrow{\kappa} : G/H_1 \rightarrow \Hom(V_1, V_2)$.
Specifically, using the decomposition $g = s(gH_1) \h{1}(g)$ of (Eq. \ref{eq:unique_decomposition}), we can define
\begin{equation}
    \kappa(g)
    = \kappa(s(gH_1)\h{1}(g))
    = \kappa(s(gH_1)) \, \rho_1(\h{1}(g))
    \equiv \overleftarrow{\kappa}(gH_1)\rho_1(\h{1}(g)),
\end{equation}
It is easy to verify that when defined in this way, $\kappa$ satisfies right $H_1$-equivariance.

We still have the left $H_2$-equivariance constraint, which translates to $\overleftarrow{\kappa}$ as follows.
For $g \in G$, $h \in H_2$ and $x \in G/H_1$,
\begin{equation}
    \begin{aligned}
        &&\kappa(h g) &\ =\ \rho_2(h) \kappa(g) \\
        \Leftrightarrow\quad
        &&\overleftarrow{\kappa}(h g H_1) \rho_1(\h{1}(h g)) &\ =\ \rho_2(h) \overleftarrow{\kappa}(g H_1) \rho_1(\h{1}(g)) \\
        \Leftrightarrow\quad
        &&\overleftarrow{\kappa}(h g H_1) &\ =\ \rho_2(h) \overleftarrow{\kappa}(g H_1) \rho_1(\h{1}(g)) \rho_1(\h{1}(h g))^{-1} \\
        \Leftrightarrow\quad
        &&\overleftarrow{\kappa}(h x) &\ =\ \rho_2(h) \overleftarrow{\kappa}(x) \rho_1(\h{1}(x, h)^{-1}),
    \end{aligned}
\end{equation}
where the last step made use of Eq. \ref{eq:h_g1g2}.

Thus, the space $\mathcal{K}_G$ of bi-equivariant, single argument kernels on $G$ is equivalent to the following space of left-equivariant kernels on $G/H_1$:
\begin{equation}
    \label{eq:def_K_C}
    \begin{aligned}
        \mathcal{K}_C = \{ \overleftarrow{\kappa} : G / H_1 \rightarrow \Hom(V_1, V_2) \, | \, & \overleftarrow{\kappa}(h_2 x) = \rho_2(h_2) \overleftarrow{\kappa}(x) \rho_1(\h{1}(x, h_2)^{-1}), \\ & \forall h_2 \in H_2, x \in G/H_1 \}
    \end{aligned}
\end{equation}
The isomorphism $\Lambda_\mathcal{K} : \mathcal{K}_C \rightarrow \mathcal{K}_G$ is defined as follows:
\begin{equation}
    \begin{aligned}
        \lbrack\Lambda_\mathcal{K} \overleftarrow{\kappa}](g) &= \overleftarrow{\kappa}(g H_1) \rho_1(\h{1}(g)), \\
        [\Lambda_\mathcal{K}^{-1} \kappa](x) &= \kappa(s(x)).
    \end{aligned}
\end{equation}
One may verify that these maps are indeed inverses, and that $\Lambda_\mathcal{K} \overleftarrow{\kappa} \in \mathcal{K}_G$ for $\overleftarrow{\kappa} \in \mathcal{K}_C$ and $\Lambda_\mathcal{K}^{-1} \kappa \in \mathcal{K}_C$ for $\kappa \in \mathcal{K}_G$.

In section \ref{sec:resolving_left_equivariance} we will resolve the left-equivariance constraint that still applies to $\mathcal{K}_C$.
But first we will continue with the $\mathcal{I}_C$ realization of $\ind$, where $\mathcal{K}_C$ will again make an appearance.

\subsection{Intertwiners for $\mathcal{I}_C$}

In this section we will study the intertwiners between two induced representations $\pi_1$ and $\pi_2$, realized on the spaces $\mathcal{I}_C^1$ and $\mathcal{I}_C^2$ (i.e. functions $G/H_1 \rightarrow V_1$ and $G/H_2 \rightarrow V_2$).
The derivations in this section will mirror those of the last section, except that we start with functions on $G/H_1$ from the start.

A general linear map $\mathcal{I}_C^1 \rightarrow \mathcal{I}_C^2$ can be written as:
\begin{equation}
    [\kappa \cdot f](x) = \int_{G/H_1} \kappa(x, y) f(y) dy
\end{equation}
where $f \in \mathcal{I}^1_G$ and $\kappa : G/H_2 \times G/H_1 \rightarrow \Hom(V_1, V_2)$.

In order for $\kappa \cdot$ to be equivariant, it must satisfy the constraint:
\begin{equation}
    \pi_2(u) [\kappa \cdot f] = \kappa \cdot [\pi_1(u) f].
\end{equation}
Expanding the left-hand side using the definition of $\pi_2$ (Eq. \ref{eq:def_ind_C}), we find
\begin{equation}
    \pi_2(u) [\kappa \cdot f] = \rho_2(\h{2}(u^{-1} x, u)) \int_{G/H_1} \kappa(u^{-1} x, y) f(y) dy
\end{equation}
For the right-hand side, we obtain
\begin{equation}
    \begin{aligned}
        \kappa \cdot [\pi_1(u) f]
        &= \int_{G/H_1} \kappa(x, y) \rho_1(\h{1}(u^{-1} y, u)) f(u^{-1} y) dy  \\
        &= \int_{G/H_1} \kappa(x, uy) \rho_1(\h{1}(y, u)) f(y) dy 
    \end{aligned}
\end{equation}
Combining the last two equations, we obtain the constraint
\begin{equation}
    \rho_2(\h{2}(u^{-1} x, u)) \; \kappa(u^{-1} x, y)
    = 
    \kappa(x, u y) \; \rho_1(\h{1}(y, u)) \\
\end{equation}
Which can be written as
\begin{equation}
    \label{eq:kernel_constraint_C}
    \kappa(x, y) = \rho_2(\h{2}(x, u))^{-1} \kappa(u x, u y) \rho_1(\h{1}(y, u))
\end{equation}

This constraint is reminiscent of the constraint for the two-argument kernel on $G$ that we found in Eq. \ref{eq:two_arg_constraint_G}, the difference being that now we get $\rho$-factors on the left and right.
As before, however, this constraint allows us to replace the two-argument kernel by a one-argument kernel.

If we take $u = s_2(x)^{-1}$, we have $u x = H_2$ (so $u \in G$ is kind of like a left inverse of $x \in G/H_2$), and $\h{2}(x, u) = e$.
Combining this with the above constraints, we obtain:
\begin{equation}
    \begin{aligned}

        \kappa(x, y) &= \rho_2(\h{2}(x, u))^{-1} \kappa(H_2, u y) \rho_1(\h{1}(y, u)) \\
        &= \kappa(H_2, u y) \rho_1(\h{1}(y, u))

    \end{aligned}
\end{equation}
This shows that we can define the two-argument kernel in terms of a one-argument kernel on $G/H_1$, defined as $\overleftarrow{\kappa}(y) = \kappa(H_2, y)$ .
The linear map $\kappa \cdot$ can then be expressed as cross-correlation with a $\rho_1$-twist:
\begin{equation}
    \label{eq:def_twisted_corr}
    \boxed{
    [\overleftarrow{\kappa} \star f](x) = \int_{G/H_1} \overleftarrow{\kappa}(s_2(x)^{-1} y) \rho_1(\h{1}(y, s_2(x)^{-1})) f(y) dy
    }
\end{equation}

The one-argument kernel is still constrained, because for $h \in H_2$, by Eq. \ref{eq:kernel_constraint_C},
\begin{equation}
    \begin{aligned}
        \overleftarrow{\kappa}(h y) &= \kappa(H_2, h y) \\
        &= \rho_2(\h{2}(H_2, h^{-1}))^{-1} \kappa(h^{-1} H_2, h^{-1}h y) \rho_1(\h{1}(h y, h^{-1})) \\
        &= \rho_2(h) \overleftarrow{\kappa}(y) \rho_1(\h{1}(y, h)^{-1}) \\
    \end{aligned}
\end{equation}
Which we recognize as the constraint for kernels in $\mathcal K_C$ that we found before in Eq. \ref{eq:def_K_C}.
Thus, we see that any intertwiner between induced representations $\mathcal{I}_C^1$ and $\mathcal{I}_C^2$ can be written as a twisted cross-correlation (Eq. \ref{eq:def_twisted_corr}) using a kernel $\overleftarrow{\kappa} \in \mathcal{K}_C$.

We note that for semidirect product groups, by Eq. \ref{eq:semidirect_h} and \ref{eq:hsx_is_e}, the $\rho_1$-twist in Eq. \ref{eq:def_twisted_corr} disappears.

\subsection{Resolving the left-equivariance constraint} \label{sec:resolving_left_equivariance}

We have seen the space $\mathcal{K}_C$ of $H_2$-equivariant kernels on $G/H_1$ appear in our analysis of both $\mathcal{I}_G$ and $\mathcal{I}_C$.
Kernels in this space have to satisfy the constraint (for $h \in H_2$):
\begin{equation}
    \overleftarrow{\kappa}(h y) = \rho_2(h) \overleftarrow{\kappa}(y) \rho_1(\h{1}(y, h)^{-1})
\end{equation}
Here we will show that this space is equivalent to the space
\begin{equation} \label{eq:def_K_D}
    \begin{aligned}
        \mathcal{K}_D = \{ \bar{\kappa} : H_2 \backslash G / H_1 \rightarrow \Hom(V_1, V_2) \, | \,
        & \bar{\kappa}(x) = \rho_2(h) \bar{\kappa}(x) \rho_1^x(h)^{-1}, \\

        & \forall x \in H_2 \backslash G / H_1, h \in H_2^{\gamma(x)H_1} \},
    \end{aligned}
\end{equation}
where we defined the representation $\rho_1^x$ of the stabilizer $H_2^{\gamma(x) H_1}$,
\begin{equation}
    \label{eq:rho1_stabilizer}
    \begin{aligned}
        \rho_1^x(h)
        &= \rho_1(\h{1}(\gamma(x)H_1, h)) \\
        &= \rho_1(\gamma(x)^{-1} h \gamma(x)),
    \end{aligned}
\end{equation}
with the section $\gamma:H_2\backslash G / H_1 \to G$ being defined as in section \ref{sec:double_cosets}.
To show the equivalence of $\mathcal{K}_C$ and $\mathcal{K}_D$, we define an ismorphism $\Omega_\mathcal{K} : \mathcal{K}_D \rightarrow \mathcal{K}_C$.
We begin by defining $\Omega_{\mathcal{K}}^{-1}$:
\begin{equation}
    \bar{\kappa}(x) = [\Omega_{\mathcal{K}}^{-1} \overleftarrow{\kappa}](x) = \overleftarrow{\kappa}(\gamma(x) H_1).
\end{equation}

We verify that for $\overleftarrow{\kappa} \in \mathcal{K}_C$ we have $\bar{\kappa} \in \mathcal{K}_D$.
Let $h \in H_2^{\gamma(x) H_1}$, then
\begin{equation}
    \begin{aligned}
        \bar{\kappa}(x) &= \overleftarrow{\kappa}(\gamma(x) H_1) \\
        &= \overleftarrow{\kappa}(h \gamma(x) H_1) \\
        &= \rho_2(h) \overleftarrow{\kappa}(\gamma(x) H_1) \rho_1(\h{1}(\gamma(x) H_1, h))^{-1} \\
        &= \rho_2(h) \bar{\kappa}(x) \rho_1^x(h)^{-1}
    \end{aligned}
\end{equation}

To define $\Omega_{\mathcal{K}}$, we use the decomposition $y = h \gamma(H_2 y) H_1$ for $y \in G/H_1$ and $h \in H_2$.
Note that $h$ may not be unique, because $H_2$ does not in general act freely on $G/H_1$.
\begin{equation} \label{eq:def_omega_K}
    \overleftarrow{\kappa}(y) = [\Omega_\mathcal{K} \bar{\kappa}](y) = [\Omega_\mathcal{K} \bar{\kappa}](h \gamma(H_2 y) H_1) = \rho_2(h) \bar{\kappa}(H_2 y) \rho_1(\h{1}(\gamma(H_2 y) H_1, h))^{-1}.
\end{equation}

We verify that for $\bar{\kappa} \in \mathcal{K}_D$ we have $\overleftarrow{\kappa} \in \mathcal{K}_C$.
\begin{equation}
    \begin{aligned}
        \overleftarrow{\kappa}(h' y)
        &= \overleftarrow{\kappa}(h' h \gamma(H_2 y) H_1) \\
        &= \rho_2(h'h) \bar{\kappa}(H_2 y) \rho_1(\h{1}(\gamma(H_2 y) H_1, h'h))^{-1} \\
        &= \rho_2(h'h) \bar{\kappa}(H_2 y) \rho_1(\h{1}(h \gamma(H_2 y) H_1, h') \h{1}(\gamma(H_2 y) H_1, h))^{-1} \\
        &= \rho_2(h') \rho_2(h) \bar{\kappa}(H_2 y) \rho_1(\h{1}(\gamma(H_2 y) H_1, h))^{-1} \rho_1(\h{1}(h \gamma(H_2 y) H_1, h'))^{-1}\\
        &= \rho_2(h') \rho_2(h) \bar{\kappa}(H_2 y) \rho_1(\h{1}(\gamma(H_2 y) H_1, h))^{-1} \rho_1(\h{1}(y, h'))^{-1}\\
        &= \rho_2(h') \overleftarrow{\kappa}(y) \rho_1(\h{1}(y, h'))^{-1}\\

    \end{aligned}
\end{equation}

We verify that $\Omega_\mathcal{K}$ and $\Omega_\mathcal{K}^{-1}$ are indeed inverses:
\begin{equation}
    \begin{aligned}
        \lbrack \Omega_\mathcal{K} [\Omega_\mathcal{K}^{-1} \overleftarrow{\kappa}]](y) 
        &=
        [\Omega_\mathcal{K} [\Omega_\mathcal{K}^{-1} \overleftarrow{\kappa}]](h \gamma(H_2 y) H_1) \\
        &= \rho_2(h) [\Omega_\mathcal{K}^{-1} \overleftarrow{\kappa}](H_2 y)\rho_1(\h{1}(\gamma(H_2 y) H_1, h))^{-1} \\
        &= \rho_2(h) \overleftarrow{\kappa}(\gamma(H_2 y) H_1) \rho_1(\h{1}(\gamma(H_2 y) H_1, h))^{-1} \\
        &= \overleftarrow{\kappa}(h \gamma(H_2 y) H_1) \\
        &= \overleftarrow{\kappa}(y).
    \end{aligned}
\end{equation}

In the other direction,
\begin{equation}
    \begin{aligned}
        \lbrack \Omega_\mathcal{K}^{-1} [\Omega_\mathcal{K} \bar{\kappa}]](x) 
        &= [\Omega_\mathcal{K} \bar{\kappa}](\gamma(x) H_1) \\
        &= [\Omega_\mathcal{K} \bar{\kappa}](\gamma(H_2 \gamma(x) H_1) H_1) \\
        &= \rho_2(e) \bar{\kappa}(H_2 \gamma(x) H_1) \rho_1(\h{1}(\gamma(H_2 \gamma(x) H_1) H_1, e))^{-1} \\
        &= \bar{\kappa}(x)
    \end{aligned}
\end{equation}

\newpage
\section{Intertwiners: Abstract Approach}
\label{sec:intertwiners_mackey}

An abstract understanding of the space of intertwiners between two induced representations can be obtained in a very direct manner by combining two fundamental results: Frobenius Reciprocity and Mackey's Lemma.
This section is based on \cite{ind_mackey2009}.

\begin{theorem}[Frobenius reciprocity]
    Let $G$ be a group, $H \leq G$ a subgroup, $(\pi, W)$ a representation of $G$, and $(\rho, V)$ a representation of $H$.
    Then

    \begin{equation}
        \Hom_G(\pi, \ind^G_H \rho) \simeq \Hom_H(\res_H^G \pi, \rho).
    \end{equation}
\end{theorem}
In words, the space of $G$-equivariant maps from $W$ to $\ind_H^G V$ is isomorphic to the space of $H$-equivariant maps from $W$ (acted by the restriction of $\pi$ to $H$) to $V$.
In abstract language, Frobenius reciprocity tells us that restriction and induction are adjoint functors.
So although the definition of $\ind$ looks a bit complicated, it is actually a very natural and unavoidable construct.

Assume now that $\pi=\ind_{H_1}^G \rho_1$ is itself induced by a representation $(\rho_1, V_1)$ of a further subgroup $H_1.$
If we apply Frobenius reciprocity to this case we find
\begin{equation}
    \Hom_G(\ind_{H_1}^G \rho_1, \ind_{H_2}^G \rho_2) \simeq \Hom_{H_2}(\res_{H_2}^G \ind_{H_1}^G \rho_1, \rho_2),
\end{equation}
where $(\rho, V)$ and $H$ were relabeled to $(\rho_2, V_2)$ and $H_2$ respectively.
This tells us that to understand the space of intertwiners between two induced representations (lhs), we need to understand the restriction of the induced representation, $\res_{H_2}^{G} \ind_{H_1}^G \rho_1$, and the $H_2$-equivariant maps from there to $(\rho_2, V_2)$.
This is where Mackey's lemma comes in.

\begin{lemma}[Mackey's lemma]
    Let $G$ be a group and $H_1, H_2$ subgroups of $G$, and let $(\rho_1, V_1)$ be a representation of $H_1$.
    Choose a section $\gamma : H_2 \backslash G / H_1 \rightarrow G$.
    Then the restriction of the induced representation decomposes as:
    \begin{equation}
        \res_{H_2}^G \ind_{H_1}^G \rho_1 \simeq \bigoplus_{x \in H_2 \backslash G / H_1} \ind_{H_2^x}^{H_2} \rho_1^x,
    \end{equation}
    where the stabilizer $H_2^x$ and its representation $\rho_1^x$ are defined as in Eq. \ref{eq:double_coset_stabilizer} and \ref{eq:rho1_stabilizer}:
    \begin{equation}
        \begin{aligned}
            H_2^x &= \gamma(x) H_1 \gamma(x)^{-1} \cap H_2, \\
            \rho_1^x(h) &= \rho_1(\gamma(x)^{-1} h \gamma(x)), \;\;\; (\forall h \in H_2^x).
        \end{aligned}
    \end{equation}
\end{lemma}
By a slight abuse of notation, we can write $\rho^x_1 = \res^{H_1}_{H_2^x}$.
It follows that
\begin{equation}
    \begin{aligned}
        \Hom_G(\ind_{H_1}^G \rho_1, \ind_{H_2}^G \rho_2) 
        &\simeq
        \Hom_{H_2}(\res_{H_2}^G \ind_{H_1}^G \rho_1, \rho_2) \\
        &\simeq
        \Hom_{H_2}\left(\bigoplus_{x \in H_2 \backslash G / H_1} \ind_{H_2^x}^{H_2} \res_{H_2^x}^{H_1}\rho_1, \rho_2 \right) \\
    \end{aligned}
\end{equation}

Finally, Mackey's formula for invariants gives a complete decomposition of the space of intertwiners.
\begin{theorem}[Mackey's formula for invariants]
    Let $G$ be a group with subgroups $H_1, H_2,$ and let $(\rho_1, V_1)$ and $(\rho_2, V_2)$ be representations of $H_1$ and $H_2$ respectively.
    Choose a section $\gamma : H_1 \backslash G / H_2 \rightarrow G$ and let $H_2^x$ and $\rho_1^x = \res_{H_2^x}^{H_1}$ be defined as before.

    Then, the space of intertwiners between the induced representations decomposes as follows:
    \begin{equation}
        \Hom_G\left(\ind_{H_1}^G\rho_1,\ind_{H_2}^G\rho_2\right)
        \simeq \bigoplus_{x \in H_2 \backslash G / H_1}

        \Hom_{H_2^x}\left(\res_{H_2^x}^{H_1} \rho_1, \res_{H_2^x}^{H_2} \rho_2 \right).
    \end{equation}
\end{theorem}
So, in order to understand the space of intertwiners, we need only need to understand the double coset space $H_2 \backslash G / H_1$ and the stabilizer $H_2^x$.

\section{Examples}
\label{sec:examples}

\subsection{Intertwiners for \SO{3}}

Let's say we want to create an equivariant network layer that maps between two fields over the sphere.
As we will see, this corresponds to taking $G = \SO{3}$ and $H_1 = H_2 = H = \SO{2}$, together with some representations $\rho_1, \rho_2$ of $\SO2$ that determine the type of field.
The theory tells us that any equivariant map can be written as a convolution with a filter $\bar{\kappa} \in \mathcal{K}_D$ (Eq. \ref{eq:def_K_D}).
The space $\mathcal{K}_D$ is defined in terms of the double coset space $H \backslash G /H$ and the stabilizer $H_2^x$, so we will proceed to compute them.

We can represent any element $g \in \SO3$ using ZYZ Euler angles as $g = Z(\alpha)Y(\beta)Z(\gamma)$, for $\alpha, \gamma \in [0, 2\pi)$ and $\beta \in [0, \pi]$.
The subgroup $H = \SO2$ corresponds to the rotations about the Z-axis, $Z(\gamma)$.

A left H-coset of the element $g = g(\alpha, \beta, \gamma)$ depends only on $\alpha, \beta$:
\begin{equation}
    \begin{aligned}
        gH  &= Z(\alpha)Y(\beta)Z(\gamma) \; \{Z(\gamma')\}_{\gamma' \in [0, 2\pi)} \\
            &= Z(\alpha)Y(\beta) \{Z(\gamma + \gamma')\}_{\gamma' \in [0, 2\pi)} \\
            &= Z(\alpha)Y(\beta) H
    \end{aligned}
\end{equation}
Hence, we can think of the coset space $G/H$ as the sphere $S^2$, parameterized by spherical coordinates $\alpha, \beta$.

Similarly, we can compute the double coset space, for which we find $H \backslash G/H = [0, \pi]$, i.e. the arc going from the north pole $\alpha, \beta = (0, 0)$ to the south pole $(\alpha, \beta) = (0, \pi)$.
Thought of as a subset of $G/H$, a double coset (i.e. point on the arc) $\beta \in H \backslash G / H$ corresponds to a circle around the Z-axis, parallel to the XY-plane, at height $\beta$.

As our section $s : G/H \rightarrow G$, we can take $s(g(\alpha, \beta, \gamma) H) = s(\alpha, \beta) = Z(\alpha)Y(\beta)$.
Note that $s(Z(\gamma) x) = Z(\gamma) s(x)$ which implies that $\h{}(x, Z(\gamma)) = e.$
For our section $\gamma : H\backslash G/H \rightarrow G$, we can take
$\gamma(H g(\alpha, \beta, \gamma) H) = \gamma(\beta) = Y(\beta)$.

So, a double coset corresponds to a point on the $\beta$ arc.
We need to determine which transformations $Z(\gamma) \in \SO2$ leave this point invariant.
For the north and south pole ($\beta=0$ and $\beta=\pi$), any rotation $Z(\gamma) \in \SO2$ leaves the point invariant, because these points lie on the Z-axis.
Hence, for these points, the stabilizer is the whole group: $H^x = \SO2$.
For other points in the $\beta$ arc, any non-zero rotation around $Z$ will move the point, so for these points the stabilizer is trivial: $H^x = \{e\}$.

Mackey's formula for invariants refers to the $H^x$-equivariant maps between two representations of $H^x$.
But since $H^x$ is trivial almost everywhere, this constraint is trivial:
\begin{equation}
    \begin{aligned}
        \Hom_G(\ind_{H}^G V_1, \ind_{H}^G V_2)
        & \simeq \bigoplus_{x \in H \backslash G / H} \Hom_{H^{x}}(\res_{H^{x}}^{H} V_1, \res_{H^x}^H V_2) \\
        & \simeq \bigoplus_{x \in H \backslash G / H} \{\Phi : V_1 \rightarrow V_1 \, | \, \Phi \rho_1^x(h) = \rho_2(h) \Phi, \;\; \forall h \in H^x \} \\
        & \simeq \bigoplus_{x \in H \backslash G / H} \{\Phi : V_1 \rightarrow V_1 \} \\

    \end{aligned}
\end{equation}
That is, because $H^x$ is trivial except at the poles, the equivariant maps are unconstrained matrix-valued kernels $\bar{\kappa} : H \backslash G / H  \rightarrow \Hom(V_1, V_2)$, or more concretely, $\bar{\kappa} : [0, \pi] \rightarrow \R^{n \times m}$ (assuming the number of input/output channels is $n, m$).
This result is in agreement with the more familiar result that in the scalar spherical convolution $f * g$, one can, without loss of generality, take $g$ to be a zonal spherical function, i.e. one that is constant in $\alpha$ \citep{Driscoll1994-ol}.

As another way of obtaining this result, we can consider the constraint on kernels $\kappa \in \mathcal{K}_C$, i.e. that for $\overleftarrow{\kappa} : G/H \rightarrow \Hom(V_1, V_2)$ we have $\overleftarrow{\kappa}(h x) = \rho_2(h) \overleftarrow{\kappa}(x) \rho_1(\h{}(x, h)^{-1})$, for all $x \in G/H, h \in H$.
If we write $h = Z(\gamma)$ and $x = (\alpha, \beta)$, then $hx = (\gamma + \alpha, \beta)$.
Using $\h{}(x, Z(\gamma)) = e$, the constraint becomes:
\begin{equation}
    \label{eq:kappa_left_constr_so3}
    \overleftarrow{\kappa}(\gamma + \alpha, \beta) = \rho_2(Z(\gamma)) \overleftarrow{\kappa}(\alpha, \beta)
\end{equation}
Therefore the value of $\overleftarrow{\kappa}$ at $(\alpha, \beta)$ can be obtained from the value at $(0, \beta)$.

Since Eq. \ref{eq:kappa_left_constr_so3} is a linear constraint, we can easily find a basis for $\overleftarrow{\kappa}$.
We would then parameterize the kernel as 
\begin{equation}
    \overleftarrow{\kappa} = \sum_i \theta_i \overleftarrow{\kappa}_i,
\end{equation}
where $\overleftarrow{\kappa}_i$ are the basis filters, and $\theta_i$ are parameters.
Alternatively, we could define $\overleftarrow{\kappa}$ in terms of a filter $\bar{\kappa}(\beta) = \overleftarrow{\kappa}(0, \beta)$, using Eq. \ref{eq:kappa_left_constr_so3}:
\begin{equation}
    \overleftarrow{\kappa}(\alpha, \beta) = \rho_2(Z(\alpha)) \bar{\kappa}(\beta).
\end{equation}
This corresponds to taking a $\bar{\kappa} \in \mathcal{K}_D$ (Eq. \ref{eq:def_K_D}) and mapping it to a kernel $\overleftarrow{\kappa} \in \mathcal{K}_C$ (Eq. \ref{eq:def_K_C}), using the isomorphism $\Omega_{\mathcal{K}}$ (Eq. \ref{eq:def_omega_K}).

Using Eq. \ref{eq:def_twisted_corr}, we can write the twisted cross-correlation as:
\begin{equation}
    \begin{aligned}

        \lbrack\overleftarrow{\kappa} \star f](x) 
        &= \int_{S^2} \overleftarrow{\kappa}(s(x)^{-1} y) \rho_1(\h{}(y, s(x)^{-1})) f(y) dy \\

    \end{aligned}
\end{equation}
Where one can fill in the above definition of $s$, $x = (\alpha, \beta)$, $y = (\alpha',\beta')$, and use the Haar measure $dy = \frac{d\alpha'}{2\pi} \frac{d\beta'}{2} \sin(\beta')$.
As far as we know, this has so far only been implemented for scalars, i.e. trivial $\rho_1$ \citep{Cohen2018-sv}.

\subsection{Intertwiners for \SE{3}}

$\SE{3} = T \rtimes H$ is a semidirect product group where $T = \R^3$ stands for the translation group and $H = \SO3$ for the rotations.
We consider $H_1 = H_2 = H$.
Then $G/H = \R^3$ and there is a unique $s$ that satisfies $s(x) \in T \; \forall x \in G/H$.
This particular $s$ implies that $\h{}(x, r) = r \; \forall r \in H$.

We will look at the space $\mathcal{K}_D$ (Eq. \ref{eq:def_K_D}),
\begin{equation}
    \begin{aligned}
        \mathcal{K}_D = \{ \bar{\kappa} : H \backslash G / H \rightarrow \Hom(V_1, V_2) \, | \,
        & \bar{\kappa}(x) = \rho_2(h) \bar{\kappa}(x) \rho_1^x(h)^{-1},
        & \forall x \in H \backslash G / H, h \in H^x \},
    \end{aligned}
\end{equation}

An element of the double coset $R \in H \backslash G/ H = [0, \infty)$ viewed as subset of $G/H= \R^3$ is a sphere of radius $R$, centered around the origin.
For the section $\gamma$ of the double coset we take the image by $s$ of the north pole of this sphere, which is the translation that brings you from the origin to the vector $(0, 0, R)$.
With this choice, the stabilizer $H^x$ is equal to $\SO2^z$, the group of rotations around the z-axis, and furthermore, $\rho_1^x(h) = \rho_1(\gamma(x)^{-1} h \gamma(x)) = \rho_1(h)$ for $h \in H^x$.

The constraint on $\mathcal{K}_D$ reduces to
\begin{equation}
    \bar{\kappa}(x) = \rho_2(h) \bar{\kappa}(x) \rho_1(h)^{-1}, \quad x \in H \backslash G / H, h \in \SO2^z
\end{equation}
which can be solved further analytically or numerically.

Eq. \ref{eq:def_omega_K} maps $\mathcal{K}_D$ to $\mathcal{K}_C$, and here it becomes:
\begin{equation}
    \overleftarrow{\kappa}(y) = \rho_2(h) \bar{\kappa}(Hy) \rho_1(h)^{-1}, \quad h \in H \text{ such that } y = h \gamma(Hy)H
\end{equation}
Alternatively, one can directly solve for the constraint $\overleftarrow{\kappa} \in \mathcal{K}_C$.

$[\overleftarrow{\kappa} \star f]$ simpifies to:
\begin{equation}
    \lbrack \overleftarrow{\kappa} \star f](x) = \int_{\R^3} \overleftarrow{\kappa}(y - x) f(y) dy
\end{equation}
That is, we just do a 3D convolution with a special constrained kernel.

$\SE3$-equivariant networks were first described and implemented by \cite{Kondor2018-fm, Thomas2018-pd}.

\section{Conclusion}

In this report we have studied the intertwiners between induced representations, which are the layers of a generalized steerable G-CNN.
We have seen how, for a wide range of groups and homogeneous spaces, the space of intertwiners can be characterized in a generic manner, as a space of constrained correlation kernels on coset or double coset spaces.
These results make the construction of steerable G-CNNs for new groups and spaces into a relatively straightforward calculation, and ensure that the most general parameterization of intertwiners is obtained.

\bibliography{refs}{}
\bibliographystyle{unsrtnat}

\end{document}